\documentclass[sigconf,nonacm]{acmart}
\usepackage[utf8]{inputenc}
\usepackage{pgfplots}
\usepackage{subcaption}
\usepackage{adjustbox}
\usepackage{mathtools}
\usepackage{graphicx}
\usepackage{url}

\usetikzlibrary{
    patterns,
    chains,
    backgrounds,
    calc,
    shadings,
    shapes.arrows,
    arrows,
    shapes.symbols,
    shadows,
    positioning,
    decorations.markings,
    backgrounds,
    arrows.meta,
    external
}
\usepackage{array}

\DeclareUnicodeCharacter{2212}{−}
\usepgfplotslibrary{groupplots,dateplot}
\usetikzlibrary{patterns,shapes.arrows}
\pgfplotsset{compat=newest}

\newif\iffinal
\finaltrue

\iffinal
  \newcommand\maxx[1]{}
  \newcommand\ian[1]{}
  \newcommand\ryan[1]{}
  \newcommand\kyle[1]{}
  \newcommand\ben[1]{}
\else
  \newcommand\maxx[1]{{\color{red}[Max: #1]}}
  \newcommand\ian[1]{{\color{blue}[Ian: #1]}}
  \newcommand\ryan[1]{{\color{magenta}[Ryan: #1]}}
  \newcommand\kyle[1]{{\color{yellow}[Kyle: #1]}}
  \newcommand\ben[1]{{\color{green}[Ben: #1]}}
\fi

\newcommand{\code}[1]{\texttt{#1}}

\begin{document}

%\title{Blob detection for Flame Spray Pyrolysis}
% \title{Leveraging High Performance Computing and Machine Learning to Optimize Flame Spray Pyrolysis Nanoparticle Synthesis}

\title{Towards Online Steering of Flame Spray Pyrolysis Nanoparticle Synthesis}

% \author{Maksim Levental, Ryan Chard, Joseph A. Libera, Kyle Chard, Jakob R. Elias, Marcus Schwarting, Marius Stan, Santanu Chaudhuri, Ian Foster}

\author{Maksim Levental}
\affiliation{\institution{University of Chicago}}
\author{Ryan Chard}
\affiliation{\institution{Argonne National Laboratory}}
\author{Joseph A. Libera}
\affiliation{\institution{Argonne National Laboratory}}
\author{Kyle Chard}
\affiliation{\institution{University of Chicago}}
\author{Aarthi Koripelly}
\affiliation{\institution{University of Chicago}}
\author{Jakob R. Elias}
\affiliation{\institution{Argonne National Laboratory}}
\author{Marcus Schwarting}
\affiliation{\institution{University of Chicago}}
\author{Ben Blaiszik}
\affiliation{\institution{University of Chicago}}
\author{Marius Stan}
\affiliation{\institution{Argonne National Laboratory}}
\author{Santanu Chaudhuri}
\affiliation{\institution{Argonne National Laboratory}}
\author{Ian Foster}
\affiliation{\institution{Argonne National Laboratory}}

\begin{abstract}
    %
    % Combustion processes driven by multiphysics across length scales, and effective optimization of these synthesis routes requires new methods to characterize the relevant phenomena at the speed of the experiment.
    Flame Spray Pyrolysis (FSP) is a manufacturing technique to mass produce engineered nanoparticles for
    applications in catalysis, energy materials, composites, and more.
    FSP instruments are highly dependent on a number of adjustable parameters,
    including fuel injection rate, fuel-oxygen mixtures,
    and temperature, which can greatly affect
    the quality, quantity, and properties of the yielded nanoparticles.
    Optimizing FSP synthesis requires monitoring,
    analyzing, characterizing, and modifying experimental conditions.
    % \maxx{The features of the nanoparticles produced by FSP are highly dependent on a number of adjustable parameters, e.g., fuel injection rate, fuel mixtures, quantity of oxygen, temperature of the flame, and more.
    % Tuning these parameters can dramatically affect the quality, quantity, size, and physical properties of the yielded nanoparticles.
    % Thus, FSP is a critical example of a synthesis process requiring rapid characterization in order to dynamically control and steer an instrument towards optimal yield.
    % An ongoing FSP experiment can be analyzed in a high frequency domain using optical instrumentation, such as Planar Laser-Induced Fluoroscopy (PLIF), and spectrometry to
    % quantify outputs.
    % Due to the rate data are collected, automated solutions are required to dynamically configure and optimize the experiment to maximize the quantity and quality of nanoparticles produced.
    % Such approaches rely on rapid characterization of the flame spray and necessitate integration with High Performance Computing resources.}
    Here, we propose a hybrid CPU-GPU Difference of Gaussians (DoG)
    method for characterizing the volume distribution
    of unburnt solution, so as to enable near-real-time optimization and
    steering of FSP experiments.
    % as indicated by PLIF imagery using a hybrid
    % CPU-GPU Difference of Gaussians (DoG) detector
    % This enables near real-time optimization and steering of FSP
    % experiments.
    Comparisons against standard implementations show our
    method to be an order of magnitude more efficient.
    This surrogate signal can be deployed as a component of an online end-to-end pipeline that maximizes the synthesis yield.
\end{abstract}

\maketitle

\section{Introduction}\label{sec:intro}
\ben{We use ``an FSP'' and ``a FSP' in various places, let's pick one}\maxx{Fixed}
% \ryan{Title?}
% \ryan{Rough intro. I think this paper is missing some story-telling around how this work
% can be used in practice. We should mention MDML and perhaps NRTO in the text.}

% Scientific instruments can produce huge quantities of data at alarmingly rapid rates,
% overwhelming experimentalists with valuable data that must be captured, analyzed,
% interpreted, cataloged, and acted upon at near the rate of arrival.
% As a result, human-managed analysis has become intractable and automated methods are required to process acquired data quickly and
% guide experiments.
% Automated approaches to optimizing and steering ongoing experiments help to realize their true value.

Flame Spray Pyrolysis (FSP) is used to manufacture
nanomaterials employed in the production of various industrial materials.
A FSP instrument rapidly produces nanoparticle precursors by combusting
liquid dissolved in an organic solvent.
Optical and spectrometry instrumentation can be used in a FSP experiment to monitor both flame and products.  
%during an experiment by using a combination of optical and spectrometry instrumentation, %Both the FSP flame and products can be monitored
%during an experiment by using a combination of optical and spectrometry instrumentation,
This provides insight into the current state and stability of the flame, quantity of fuel successfully combusted, and quantity and quality of materials produced.
To avoid fuel wastage and to maximize outputs, data must not only be collected but also analyzed and acted upon continuously during an experiment.

A key technique for monitoring and determining a FSP experiment's effectiveness
is quantifying and classifying the fuel that is not successfully combusted.
Planar Laser-Induced Fluoroscopy (PLIF) and optical imagery of the flame can
be used to make informed decisions about an experiment and guide the fuel mixtures, fuel-oxygen rates,
and various pressures.
Fuel wastage can be approximated by detecting and quantifying the droplets of fuel that have not been successfully aerosolized (thus, uncombusted).
Blob detection algorithms provide an effective means for both detecting fuel droplets and determining their sizes.
Blob detection is the identification and characterization of regions in an image that are distinct from adjacent regions and for which certain properties are either constant or approximately constant.
Note, however, that the volume and rate of the data generated by FSP instruments requires both automation and high-performance computing (HPC); the primary latency constraint is imposed by the collection frequency of the imaging sensor (ranging from 10Hz-100Hz).
%to meet the data processing needs of an ongoing experiment.

\ian{I find myself a bit unclear here as to the specific goals: is it understanding what is happening (e.g., quantifying unburnt fuel) or controlling activities? The former for the latter? What are the data rates, response rates?}
\maxx{I think the trouble is that while near term this technique is useful for just characterization, aspirationally we imagine it could be used to steer. Data rates are roughly catalogued in various places (10Hz sensor, less than a millisecond network latency, approx 100ms to run this image processing technique). I guess you're suggesting we should put it front and center since it's part of our pitch?}\maxx{Fixed}

Argonne National Laboratory's Combustion Synthesis Research Facility (CSRF) operates a FSP experiment that produces silica, metallic, oxide and alloy powders or particulate films.
However, standard processing methods on CPUs far exceeds FSP edge computing capacity, i.e. performing image analysis on the FSP device is infeasible.
% \ian{What exceeds capacity? You mean you want more computing to reduce the 8s to 0.1 sec, say? That could be clearer.}
% \maxx{Not sure what this means exactly but I would guess your interpretation is correct and that this is an allusion to being compute bound "at the end"}\maxx{Fixed}
To address this need for rapid, online analysis we present a hybrid CPU-GPU method to rapidly evaluate the volume distribution of unburnt fuel.
%enable near real-time optimization and
%steering of FSP experiments.

\begin{figure}
    \centering
    \scalebox{0.75}{\includegraphics{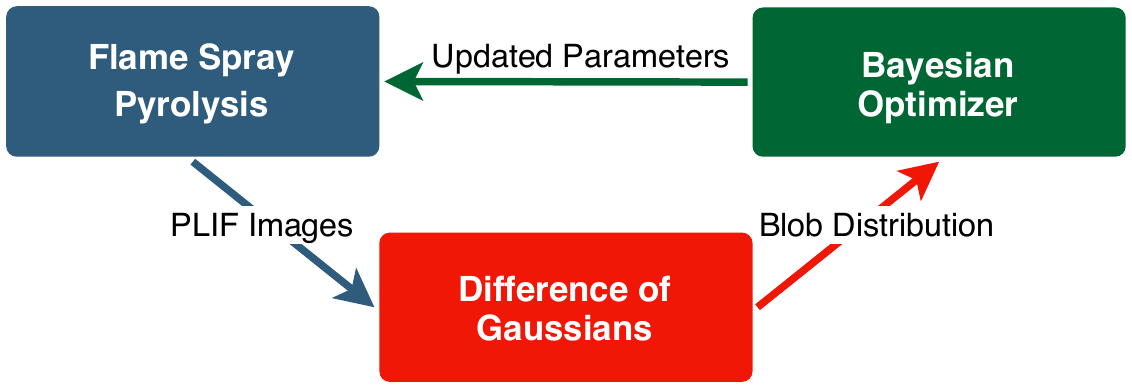}}
    \caption{End-to-end system for optimizing FSP synthesis.}
    \label{fig:block_diagram}
\end{figure}

Surrogate models can be employed to steer experiments by producing an inference for a quantity of interest faster than physical measurements~\cite{liu2019tomogan}.
In particular, machine learning surrogate models have been used in materials science, with varying degrees of success~\cite{guan2019ptychonet}.
Here we aim to drive a Bayesian hyper-parameter optimizer (BO) that optimizes the various parameters of CSRF's FSP experiment.
The combination of FSP, our technique, and the BO, define an "end-to-end" system for optimizing FSP synthesis (see figure~\ref{fig:block_diagram}).
We integrate our solution into the Manufacturing Data and Machine Learning platform~\cite{elias2020manufacturing}, enabling near-real-time optimization and steering of FSP experiments through online analysis.

The remainder of this paper is as follows: Section \ref{sec:science} discusses the materials science of FSP, Section \ref{sec:blobdetection} discusses blob detection and derives some important mathematical relationships, Section \ref{sec:implementation} covers practical considerations and our improvements to blob detection, Section \ref{sec:evaluation} compares our improved algorithms against standard implementations,
Section~\ref{sec:related} discusses related work, and Section \ref{sec:conclusion} concludes.

\section{Flame Spray Pyrolysis}\label{sec:science}

\begin{figure}[!htbp]
        \centering
        \includegraphics[width=.8\linewidth,keepaspectratio]{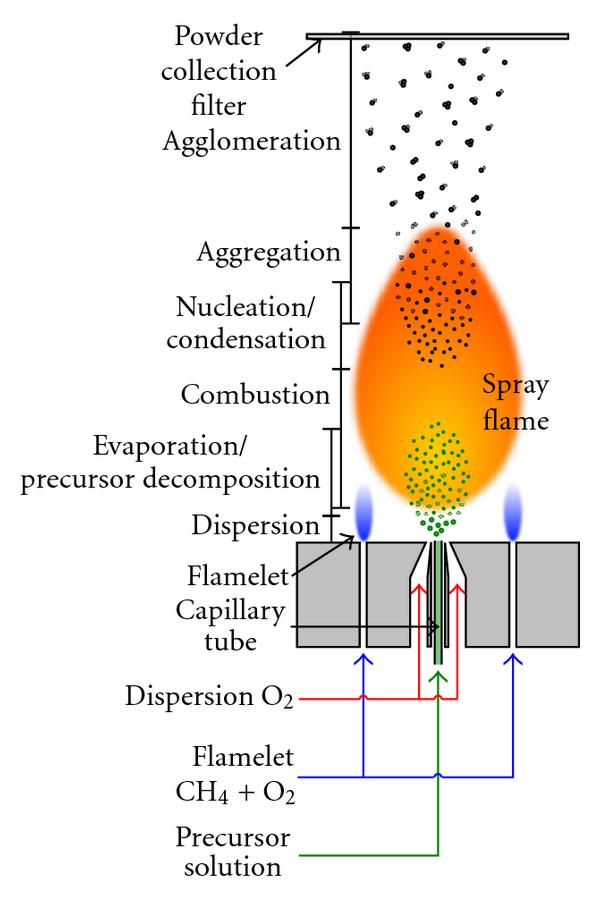}
        \caption{Schematic of the flame spray pyrolysis system and the particle formation process. Reprinted from~\cite{transition}.}
        \label{fig:fsp}
\end{figure}

Flame aerosol synthesis (FAS) is the formation of fine particles from flame gases~\cite{PRATSINIS1998197}.
Pyrolysis (also thermolysis) is the separation of materials at elevated temperatures in an inert environment (that is, without oxidation).
Flame Spray Pyrolysis (FSP) is a type of FAS where the precursor for the combustion reaction is a high heat of combustion liquid dissolved in an organic solvent~\cite{SOKOLOWSKI1977219} (
see \figurename~\ref{fig:fsp}).
An early example of industrial manufacturing employing FSP was in the production of carbon black, used, for example, as a pigment and a reinforcing material in rubber products.
% Carbon black synthesis proceeds from the reaction of a hydrocarbon fuel with limited combustion air to enthalpically pyrolyze the remaining solution~\cite{donnet1993carbon}. \ian{If short of space, this info may not be needed.}

% \begin{figure}[!htbp]
%     \centering
%     \begin{subfigure}{.45\textwidth}
%         \centering
%         \includegraphics[width=.8\linewidth,keepaspectratio]{figures/fsp/fsp.jpg}
%         \caption{Schematic of the flame spray pyrolysis system and the particle formation process. Reprinted with permission from~\cite{transition}.}
%         \label{fig:fspdiagram}
%     \end{subfigure}
    
%     \vspace{1.5em}
    
%     \begin{subfigure}{.45\textwidth}
%         \centering
%         \includegraphics[width=.8\linewidth,keepaspectratio]{figures/fsp/fsp_fire.jpg}
%         \caption{FSP at the Argonne Combustion Synthesis Research Facility.}
%         \label{fig:flamethrower}
%     \end{subfigure}
%     \caption{Flame Spray Pyrolysis}
%     \label{fig:fsp}
% \end{figure}

%Coupling high-frequency instrument parameters (including sheath O\textsubscript{2}, pilot CH\textsubscript{4} and solution flow rates, solute concentration, and air gap) with low-frequency physical measurements makes optimizing FSP reaction yield challenging.
\ian{I am not sure where the text that is commented out after this sentence comes from, but it is much clearer!}
\maxx{I agree since I wrote it! Not sure why someone changed.}\maxx{Fixed}
Optimizing FSP reaction yield is challenging; there are many instrument parameters (sheath O\textsubscript{2}, pilot CH\textsubscript{4} and solution flow rates, solute concentration, and air gap) and salient physical measurements often have low resolution in time.
Reaction yield can be inferred from the FSP droplet size distribution, with smaller droplets implying a more complete combustion.
Unfortunately, the characterization of droplet distribution via, for example, Scanning Mobility Particle Sizer Spectrometry (SMPSS) has an analysis period\ian{what is ``period''?}\maxx{fixed} of roughly ten seconds~\cite{smpss},\ian{why do we describe SMPSS if it is no good?}
\maxx{I think it functions as a foil for our heroic technique :) and it is in fact what Joe currently uses.}
while other measurement techniques, such as Raman spectroscopy, require even longer collection periods.
These measurement latencies prevent the use of such characterization techniques for adaptive control of the experiment and hence optimization of experiment parameters.

\begin{figure}[!htbp]
        \centering
       \includegraphics[width=.9\columnwidth]{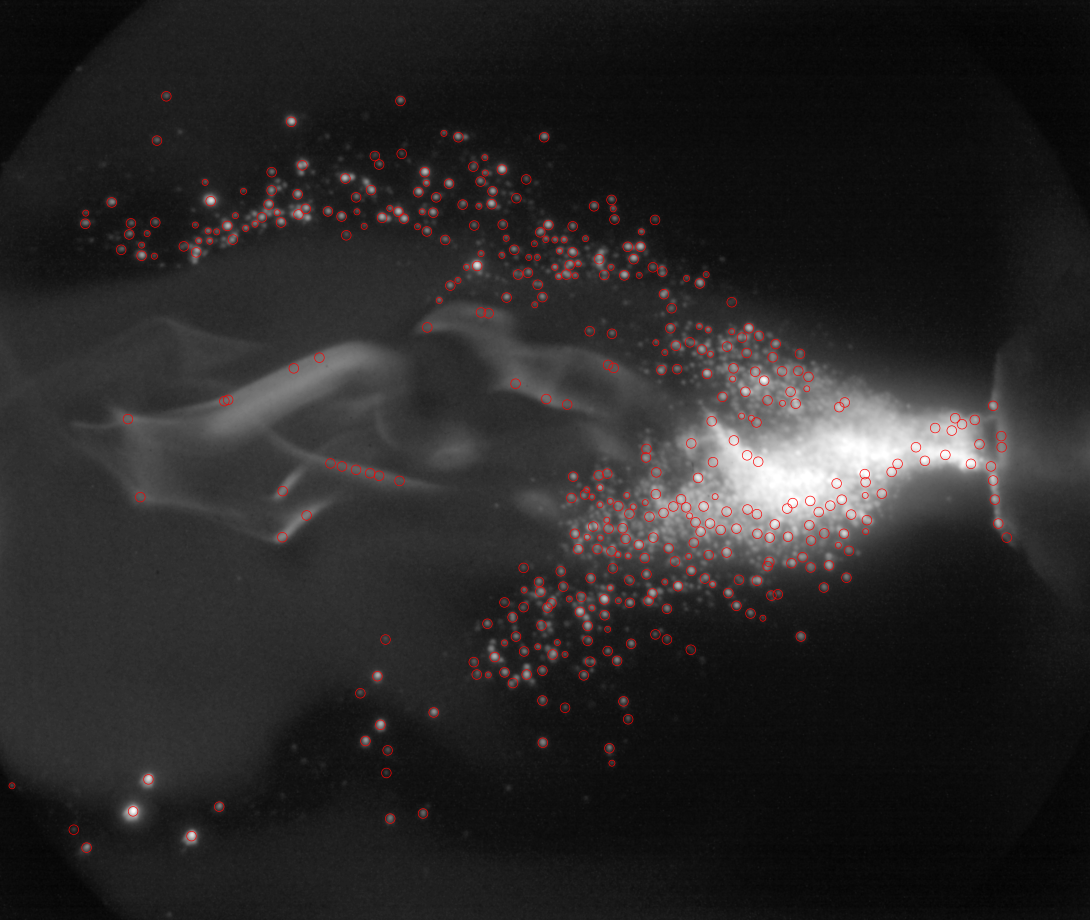} 
    \caption{FSP imagery and droplet detected by using our method. Note that in areas of high-turbulence (e.g. directly in front of the nozzle) overlapping droplets are extremely difficult to detect.}
    \label{fig:plif}
\end{figure}

In this work, droplet imaging is performed by using a iSTAR sCMOS camera oriented perpendicular to a 285nm laser light sheet that bisects the flame along its cylindrical axis. 
Imagery is collected at 10Hz, with each image capturing the droplet distribution from a 10ns exposure due to the laser pulse. 
The images are formed by Mie scattering from the surfaces of the spherical droplets (see \figurename~\ref{fig:plif}). 
The image size is not calibrated to droplet size, and so we use relative size information to assess the atomization efficiency of the FSP.

% \begin{figure}[!htbp]
%     \centering
%     \begin{subfigure}{.45\textwidth}
%         \centering
%       \includegraphics[width=.9\textwidth]{figures/plif/plif.png} 
%         \caption{PLIF instrument sample. Diffuse regions are indeed induced fluorescence, while brighter regions are the result of Mie scattering by poorly atomized fuel droplets.}
%         \label{fig:plif1}
%     \end{subfigure}
    
%     \vspace*{1.5em}
    
%     \begin{subfigure}{.45\textwidth}
%         \centering
%       \includegraphics[width=.9\textwidth]{figures/plif/plif_bubbles.png} 
%         \caption{Blobs/droplets detected in PLIF}
%         \label{fig:plif2}
%     \end{subfigure}
%     \caption{PLIF imagery and resulting droplet detection using our method.}
%     \label{fig:plif}
% \end{figure}

\ben{This figure looks impressive from far away, but when you zoom in you can see that a lot of small blobs are being missed. It seems these small blobs are very important to understanding and optimizing the process. I wonder if anyone has thoughts on how to catch these without adding too much noise.}
\maxx{You can't - it's basically the ol' Bias - Variance tradeoff problem - try pick up smaller droplets and you pick up noise, try to reduce sensitivity to noise and you exclude smaller droplets. The only real solution is a model of the generating process (i.e. a biased estimator).}

\section{Blob detection background}\label{sec:blobdetection}

\ben{Can we call this a segmentation problem? It might make it clear to the reader}\maxx{Indeed segmentation is another way to solve this problem but we're not segmenting here since we're not returning pixel level labels.}

Blob detectors operate on scale-space representations~\cite{Lindeberg1998}; define the scale-space representation $L(x,y,\sigma)$ of an image $I(x,y)$ to be the convolution of that image with a mean zero Gaussian kernel $g(x,y,\sigma)$, i.e.:
$$
L(x,y,\sigma) \coloneqq g(x,y,\sigma) * I(x,y),
$$
where $\sigma$ is the standard deviation of the Gaussian and determines the \textit{scale} of $L(x,y,\sigma)$.
Scale-space blob detectors are also parameterized by the scale $\sigma$, such that their response is extremal when simultaneously in the vicinity of blobs and when $\sqrt{2}\sigma = r$ corresponds closely to the characteristic radius of the blob.
In particular, they are \textit{locally} extremal in space and \textit{globally} extremal in scale, since at any given scale there might be many blobs in an image but each blob has a particular scale.
Conventionally, scale-space blob detectors are strongly positive for dark blobs on light backgrounds and strongly negative for bright blobs on dark backgrounds.
Two such operators are:
\begin{itemize}
    \item \textit{Scale-normalized trace of the Hessian}:
    $$
        \operatorname{trace}(H L) \coloneqq \sigma^2\left(\frac{\partial^2 L}{\partial x^2} + \frac{\partial^2 L}{\partial y^2}\right)
    $$
    Equivalently $\operatorname{trace}(H L) \coloneqq \sigma^2\Delta L$, where $\Delta$ is the Laplacian.
    \item \textit{Scale-normalized Determinant of Hessian} (DoH):
    $$
        \operatorname{det} (H L)  \coloneqq \sigma^4 \left(\frac{\partial^2 }{\partial x^2}\frac{\partial^2 L}{\partial y^2}-\left(\frac{\partial^2 L}{\partial x \partial y}\right)^2\right)
    $$
\end{itemize}
Note that by the associativity of convolution,
$$
\sigma^2\Delta L = \sigma^2 \Delta \left(g * I\right) = \left( \sigma^2\Delta g \right) * I
$$
Consequently, $\sigma^2\Delta g$ is called the \textit{scale-normalized Laplacian of Gaussians} (LoG) operator/blob detector.

\begin{figure}[!htbp]
    \centering
    \begin{subfigure}{.45\textwidth}
        \centering
        \begin{tikzpicture}
            \node[anchor=south west,inner sep=0] (image) at (0,0) {\includegraphics[width=.9\textwidth]{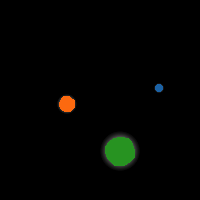}};
            \begin{scope}[x={(image.south east)},y={(image.north west)}]
                \draw[help lines,xstep=.02,ystep=.02, color=white] (0,0) grid (1,1);
                \foreach \x in {0,20,...,200} { \node [anchor=north] at (\x/200,0) {\x}; }
                \foreach \y in {0,20,...,200} { \node [anchor=east] at (0,\y/200) {\y}; }
            \end{scope}
        \end{tikzpicture}
        \caption{Blobs at different scales}
        \label{fig:scalespace-blobs}
    \end{subfigure}
    
    \vspace*{1.5em}
    
    \begin{subfigure}{.45\textwidth}
        \centering
        \begin{adjustbox}{width=1\textwidth}

        % \begin{tikzpicture}[trim axis left, trim axis right]
        \begin{tikzpicture}
            \definecolor{color0}{rgb}{0.12156862745098,0.466666666666667,0.705882352941177}
            \definecolor{color1}{rgb}{1,0.498039215686275,0.0549019607843137}
            \definecolor{color2}{rgb}{0.172549019607843,0.627450980392157,0.172549019607843}

            \begin{axis}[
                ymajorticks=false,
                % legend cell align={left},
                % legend style={fill opacity=0.8, draw opacity=1, text opacity=1, draw=white!80!black},
                % tick align=outside,
                tick pos=left,
                x grid style={white!69.0196078431373!black},
                xlabel={$\sigma$},
                ylabel={response magnitude},
                xmin=-1.95, xmax=40.95,
                xtick style={color=black},
                ymin=-0.0407895042066812, ymax=0.609148369751114
            ]
                \addplot [semithick, color0]
                table {%
                    0 0.234704852104187
                    1 0.409044728111476
                    2 0.392784005180001
                    3 0.320309805860743
                    4 0.252128031179309
                    5 0.198999357875437
                    6 0.159241870027035
                    7 0.129501174176112
                    8 0.106978711895645
                    9 0.0896482763625681
                    10 0.0760916430270299
                    11 0.0653108284436166
                    12 0.0565924091869965
                    13 0.0494045759330038
                    14 0.0433273670170456
                    15 0.0380328239756636
                    16 0.0332633741758764
                    17 0.0288272484461777
                    18 0.0246046733704861
                    19 0.0205236744898139
                    20 0.016564721846953
                    21 0.0127419390133582
                    22 0.00908892343868502
                    23 0.00565108978713397
                    24 0.00247324878815562
                    25 -0.000405492435675114
                    26 -0.00295399863040075
                    27 -0.00515349493653048
                    28 -0.00699850041419268
                    29 -0.00848805560555775
                    30 -0.00963601394905709
                    31 -0.0104608639079379
                    32 -0.0109865471161902
                    33 -0.0112379786476959
                    34 -0.0112468735722359
                    35 -0.0110431151173543
                    36 -0.0106529415748082
                    37 -0.010109300048789
                    38 -0.00943600871716626
                    39 -0.00865803239517845
                };
                % \addlegendentry{a}
                \addplot [semithick, color1]
                table {%
                    0 0.0642142295837402
                    1 0.243363718651235
                    2 0.426214355453849
                    3 0.520507374033332
                    4 0.529645072966814
                    5 0.493415249884129
                    6 0.441193641237915
                    7 0.387599218832329
                    8 0.338489550054073
                    9 0.295589500516653
                    10 0.258878936246037
                    11 0.227681832304224
                    12 0.201193362027407
                    13 0.17856799445115
                    14 0.159050338687375
                    15 0.142013736348599
                    16 0.126941575556994
                    17 0.113430612294469
                    18 0.101204592008144
                    19 0.0900742974481545
                    20 0.0799156920053065
                    21 0.0706578257167712
                    22 0.062270294977352
                    23 0.0547217297437601
                    24 0.0479846890456974
                    25 0.0420392758096568
                    26 0.036849063411355
                    27 0.0323662780830637
                    28 0.0285470149107277
                    29 0.0253415019693784
                    30 0.0226852282648906
                    31 0.0205244055471849
                    32 0.0187993113696575
                    33 0.0174611804482993
                    34 0.0164496337319724
                    35 0.0157204577641096
                    36 0.0152261331956834
                    37 0.014933664330747
                    38 0.0148026300035417
                    39 0.0147996754373889
                };
                % \addlegendentry{b}
                \addplot [semithick, color2]
                table {%
                    0 0.00219446420669556
                    1 0.0330440223217011
                    2 0.109310063645244
                    3 0.21562477722764
                    4 0.331119357347488
                    5 0.434934229031205
                    6 0.512637915983796
                    7 0.559760890081525
                    8 0.579605739116669
                    9 0.57866966214031
                    10 0.56359624620527
                    11 0.539819854851812
                    12 0.511338163465261
                    13 0.48081783445552
                    14 0.449999327473342
                    15 0.41995651954785
                    16 0.391237385571003
                    17 0.364165154881775
                    18 0.338852920066565
                    19 0.315311680166051
                    20 0.293505977094173
                    21 0.273355115568265
                    22 0.254791703633964
                    23 0.237722261678427
                    24 0.222045799903572
                    25 0.207676401827484
                    26 0.194521913044155
                    27 0.182486547040753
                    28 0.171491763759404
                    29 0.161423476613127
                    30 0.152235418837517
                    31 0.143818747554906
                    32 0.136114218831062
                    33 0.129052671068348
                    34 0.122570512760431
                    35 0.116608411911875
                    36 0.111117036882788
                    37 0.106055232044309
                    38 0.101370166577399
                    39 0.0970330754644237
                };
                % \addlegendentry{c}
            \end{axis}

        \end{tikzpicture}
        \end{adjustbox}
        \caption{Convolution kernel magnitude responses for different scale $\sigma$ values. Plot colors correspond to blob colors.}
        \label{fig:scalespace-plots}
    \end{subfigure}
    \caption{Scale-space blob detectors. Each response is measured at the center of the corresponding blob; notice that magnitude responses are maximal at $\sigma = r/\sqrt{2}$ corresponding to the radius of detected blob.}
    \label{fig:scalescapedetector}
\end{figure}
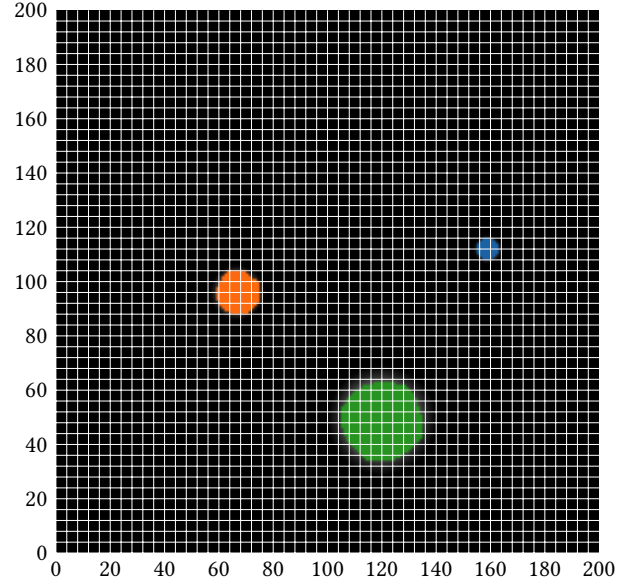
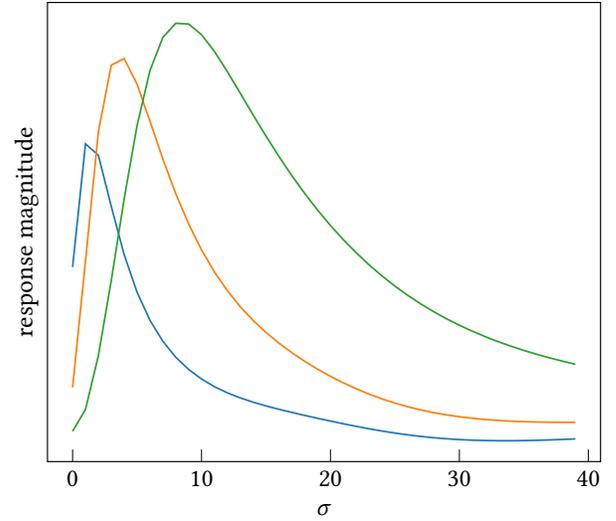

DoH and LoG are both nonlinear (due to dependence on derivatives) and compute intensive (directly a function of the dimensions of the image), but in fact LoG can be approximated.
Since $g$ is the Green's function for the heat equation: % we have
\begin{align*}
   \sigma \Delta g &= \frac{\partial g}{\partial \sigma} \\
    &\approx \frac{g(x,y,\sigma + \delta \sigma) - g(x,y,\sigma)}{\delta \sigma}
\end{align*}
Hence, we see that for small, fixed, $\delta \sigma$
$$
\sigma^2 \Delta g \approx \sigma \times [g(x,y,\sigma + \delta \sigma) - g(x,y,\sigma)]
$$
up to a constant.
This approximation is called the \textit{Difference of Gaussians} (DoG).
In practice, one chooses $k$ standard deviations such that the $\sigma_i$ densely sample the range of possible blob radii, through the relation  $\sqrt{2}\sigma_i = r_i$.
For example, \figurename~\ref{fig:scalescapedetector} schematically demonstrates the response of a DoG detector that is responsive to approximately 40 blob radii, applied to three blobs.

While LoG is the most accurate detector, it is also slowest as it requires numerical approximations to second derivatives across the entire image.
In practice, this amounts to convolving the image with second order \textit{edge filters}, which have large sizes and thereby incur a large number of multiply–accumulate (MAC) operations.
DoG, being an approximation, is faster than LoG, with similar accuracy.
Hence, here we focus on LoG and derivations thereof.

\section{Blob Detector Implementation}\label{sec:implementation}
We abstractly describe a standard implementation of a blob detector that employs DoG operators, and then discuss optimizations.

A first, optional, step is to preprocess the images; a sample image (one for which we wish to detect blobs) is smoothed and then contrast stretched (with 0.35\% saturation) to the range $[0,1]$.
This preprocessing is primarily to make thresholding more robust (necessary for us because our imaging system has dynamic range and normalizes anew for every collection).
Since DoG is an approximation of LoG, a sample image is first filtered by a set of mean zero Gaussian kernels $\{g(x,y,\sigma_i)\}$ in order to produce the set of scale-space representations $\{L(x,y,\sigma_i)\}$.
The quantity and standard deviations of these kernels is determined by three hyperparameters:
\begin{itemize}
    \item \code{n\_bin} --- one less than the number of Gaussian kernels. This hyperparameter ultimately determines how many different radii the detector can recognize.
    \item \code{min\_sigma} --- proportional to the minimum blob radius recognized by the detector.
    \item \code{max\_sigma} --- proportional to the maximum blob radius recognized by the detector.
\end{itemize}
Therefore, define $\delta\sigma \coloneqq (\mathtt{max\_sigma} - \mathtt{min\_sigma})/\mathtt{n\_bin}$ and
\begin{equation}
    \sigma_i \coloneqq \mathtt{min\_sigma} + (i-1) \times \delta\sigma
\end{equation}
for $i=1, \dots, (\mathtt{n\_bin}+1)$.
This produces an arithmetic progression of kernel standard deviations such that $\sigma_1 = \mathtt{min\_sigma}$ and $\sigma_{\mathtt{n\_bin}} = \mathtt{max\_sigma}$.
Once an image has been filtered by the set of mean zero Gaussian kernels, we take adjacent pairwise differences
\begin{equation}
    \Delta L_i \coloneqq L(x,y,\sigma_{i+1})-L(x,y,\sigma_i)
\end{equation}
for $i=1, \dots, \mathtt{n\_bin}$,
and define
\begin{equation}
    \operatorname{DoG}(x,y,\sigma_i) \coloneqq \sigma_i \times \Delta L_i
\end{equation}
We then search for local maxima
\begin{equation*}
    \{(\hat{x}_j, \hat{y}_j, \hat{\sigma}_j)\} \coloneqq \operatorname*{argmaxlocal}_{x,y, \sigma_i} \operatorname{DoG}(x,y,\sigma_i)
\end{equation*}
Such local maxima can be computed in various ways (for example, zero crossings of numerically computed derivatives) and completely characterize circular blob candidates with centers $(\hat{x}_j, \hat{y}_j)$ and radii $\hat{r}_j = \sqrt{2}\hat{\sigma}_j$.
Finally blobs that overlap in excess of some predetermined, normalized, threshold can be pruned or coalesced.
Upon completion, the collection of blob coordinates and radii $\{(\hat{x}_j, \hat{y}_j, \hat{\sigma}_j)\}$ can be transformed further (for example, histogramming to produce a volume distribution).
% See figure~\ref{fig:dogconv} for a schematic example with $\mathtt{n\_bin} = 5$.

An immediate, algorithmic, optimization is with respect to detecting extrema.
We use a simple heuristic to identify local extrema: for smooth functions, values that are equal to the extremum in a fixed neighborhood are in fact extrema.
% (see figure~\ref{fig:maxfilter})
We implement this heuristic by applying an $n \times n$ maximum filter $\operatorname{Max}(n,n)$ (where $n$ corresponds to the neighborhood scale), to the data and comparing:
$$
    \operatorname*{argmaxlocal}_{x,y} I \equiv \{x,y \,| (\operatorname{Max}(n,n)*I - I) = 0 \}
$$
Second, note that since $\operatorname{DoG}(x,y,\sigma_i)$, in principle, is extremal for exactly one $\sigma_i$ (that which corresponds to the blob radius) we have
$$
\operatorname*{argmaxlocal}_{x,y, \sigma_i} \operatorname{DoG}(x,y,\sigma_i) = \operatorname*{argmaxlocal}_{x,y} \operatorname*{argmax}_{\sigma_i} \operatorname{DoG}(x,y,\sigma_i)
$$
Thus, we can identify local maxima of $\operatorname{DoG}(x,y,\sigma_i)$ by the aforementioned heuristic, by applying a 3D maximum filter to the collection $\operatorname{Max}(n,n,n)$ to $\{\operatorname{DoG}(x,y,\sigma_i)\}$.
Typically, neighborhood size is set to three.
This has the twofold effect of comparing scales with only immediately adjacent neighboring scales and setting the minimum blob proximity to two pixels.

Further optimizations are implementation dependent.
A reference sequential (CPU) implementation of the DoG blob detector is available in the popular Python image processing package scikit-image ~\cite{scikit-image}.
For our use case, near real-time optimization, this reference implementation was not performant for dense volume distributions (see Section~\ref{sec:evaluation}).
Thus, we developed a de novo implementation; the happily parallel structure of the algorithm naturally suggests itself to GPU implementation.

\subsection{GPU Implementation}

We use the PyTorch~\cite{NEURIPS2019_9015} package due its GPU primitives and ergonomic interface.
Our code and documentation are available~\cite{merf-fsp-github}.
As we use PyTorch only for its GPU primitives, rather than for model training, our implementation consists of three units with ``frozen'' weights and a blob pruning subroutine.
%(see \figurename~\ref{fig:gpudog}).

% \input{figures/doggpu}

We adopt the PyTorch $\mathtt{C_{out}} \times \mathtt{H} \times \mathtt{W}$ convention for convolution kernels and restrict ourselves to grayscale images.
Hence, the first unit in our architecture is a \code{(n\_bin+1) $\times$ max\_width $\times$ max\_width} 2D convolution, where \code{max\_width} is proportional to the maximum sigma of the Gaussians (discussed presently).
Note that the discretization mesh for these kernels must be large enough to sample far into the tails of the Gaussians.
To this end, we parameterize this sampling $\mathtt{radius}_i$ as $\mathtt{radius}_i \coloneqq   \mathtt{truncate} \times \sigma_i$
and set $\mathtt{truncate} = 5$ by default (in effect sampling each Gaussian out to five standard deviations).
The width of each kernel is then $\mathtt{width}_i \coloneqq   2\times\mathtt{radius}_i + 1$ in order that the Gaussian is centered relative to the kernel.
% In order that this
To compile this convolution unit to a single CUDA kernel, as opposed to a sequence of kernels, we pad with zeros all of the convolution kernels (except the largest) to the same width as the largest such kernel.
That is to say
\ben{This sentence feels too complex, we should break it up for ease of reading}\maxx{Fixed}
$$
\mathtt{width}_i = \mathtt{max\_width} \coloneqq   \max_j\{\mathtt{width}_j\}
$$

The second unit is essentially a linear layer that performs the adjacent differencing and scaling discussed in Section~\ref{sec:implementation}.
The maximum filtering heuristic is implemented using a $3 \times 3 \times 3$ 3D maximum pooling layer with stride equal to 1 (so that the DoG response isn't collapsed).
Blobs
% candidates
are then identified and low confidence blobs (DoG response lower than some threshold, by default $0.1$) are rejected
$$
    \operatorname{Thresh}(x,y,\sigma) \coloneqq \begin{cases}
    1 \text{ if } \left(\operatorname{MaxPool3D} * D - D = 0\right) \bigwedge (D > 0.1) \\
    0 \text{ otherwise}
    \end{cases}
$$
where $D \coloneqq D(x,y,\sigma)$.
Once true blobs are identified, overlapping blobs can be pruned according to an overlap criterion; pruning proceeds by finding pairwise intersections amongst blobs and coalescing blobs (by averaging the radius) whose normalized overlap exceeds a threshold (by default $0.5$).
Blobs are then histogrammed, with bin centers are equal to $\{\hat{r}_i\}$.
Currently pruning and histogramming are both performed on the CPU but could be moved to the GPU as well~\cite{Johnson_2019, gpuhistogram}.

Surprisingly, for the naive PyTorch implementation of DoG there is a strong dependence on \code{max\_sigma}. This turns out to be due to how convolutions are implemented in PyTorch.
Further refining the GPU implementation by implementing FFT convolutions sacrifices (slightly) independence with respect to \code{n\_bin}, but renders the detector constant with respect to \code{max\_sigma} and dramatically improves performance (see Section~\ref{sec:evaluation}).

% We also explore model parallelism.
% This is done by partitioning up the first Conv2D layer of the network across $n=2,3,4$ GPUs, mapping the imaging across the GPUs, and then reducing by concatenation prior to performing differencing-rescaling.
% Our results suggest that scaling to many more GPUs across multiple nodes is feasible.
% \maxx{what else to say here?}

\section{Evaluation}\label{sec:evaluation}

We compare our GPU implementation of DoG against the standard implementation found in scikit-image.
Our test platform is described in Table~\ref{tab:test}.
Both detectors were tested on simulated data generated by a process that emulates the FSP experiment; a graphics rendering engine renders 100 spheres and then Poisson and Gaussian noise~\cite{Cesarelli2013} is added to the rendered image.
The images have a resolution of 1000 $\times$ 1000 pixels.
We turn off overlap pruning as it is the same across both methods (performed on the CPU).
In order to perform an "apples to apples" comparison we exclude disk I/O and Host-GPU copy times from the charted measurements; we discuss the issue of Host-GPU copy time as potentially an inherent disadvantage of our GPU implementation in the forthcoming.

% Precision and recall, according to PASCAL VOC 2012~\cite{pascal-voc-2012} conventions, for 100 such samples is presented in \figurename~\ref{fig:precisionrecall}; we compare precision and recall for both the CPU and GPU implementations.
We compare precision and recall, according to PASCAL VOC 2012~\cite{pascal-voc-2012} conventions, for both the CPU and GPU implementations.
The precision and recall differences have means $-3.2 \times 10^{-4}, 3.5 \times 10^{-3}$ and standard deviations $4.0\times 10^{-6}, 2.0\times 10^{-4}$ respectively.
This shows that for a strong majority of samples, the precision and recall for a given sample are exactly equal for both implementations; on rare occasions there is a difference in recall due to the padding of the Gaussian kernels for the GPU implementation.

\begin{table}
\caption{Test platform}

\vspace{-2ex}

\centering
\begin{tabular}[t]{p{0.15\linewidth}p{0.75\linewidth}}
\hline
CPU & Intel Xeon Gold 6230 CPU @ 2.10GHz \\
GPU & Tesla V100-PCIE-16GB \\
HD & HPE 800GB SAS 12G Mixed Use SFF \\
RAM & 384GB \\
Software & scikit-image-0.16.2, PyTorch-1.5.0, CUDA-10.2, NVIDIA-440.33.01\\
\hline
\end{tabular}
\label{tab:test}
\end{table}%

The performance improvement of the straightforward translation of the standard algorithm to PyTorch is readily apparent.
The PyTorch implementation is approximately constant in \code{n\_bin} (i.e. the number of Gaussian filters) owing to the inherent parallelism of GPU compute (see \figurename~\ref{fig:gpu_vs_cpu}) while the CPU implementation scales linearly in \code{n\_bin} (owing to the sequential nature of execution on the CPU).
We see a discontinuous increase in runtime for the standard GPU implementation at $\mathtt{n\_bin} = 32$ due to CUDA's internal optimizer's choice of convolution strategy; setting a fixed strategy is currently not exposed by the CUDA API.
Naturally, both implementations scale linearly in the \code{max\_sigma} (i.e. the widths of the Gaussian filters).
On the other hand, CPU as compared with our bespoke FFT convolution implementation is even more encouraging; FFT convolutions make the GPU implementation even faster.
They are weakly dependent (approximately log-linear) on \code{n\_bin} but constant with respect to \code{max\_sigma} (see \figurename~\ref{fig:gpu_vs_gpu}).
In fact, we see that for $\mathtt{max\_sigma} > 2$ and $\mathtt{n\_bin} > 10$ it already makes sense to use FFT convolution instead of the standard PyTorch convolution.

\begin{table}
\caption{Various times of merit}

\vspace{-2ex}

\centering
\begin{tabular}[t]{p{0.45\linewidth}p{0.45\linewidth}}
\hline
Host-GPU copy & 100$\mu$s--200 $\mu$s \\
Disk I/O & ${\sim}20$ms \\
Preprocessing & ${\sim}90$ms \\
Contrast stretching & ${\sim}70$ms \\
\hline
\end{tabular}
\label{tab:iotimes}
\end{table}%

Some discussion of I/O and Host-GPU and GPU-Host copy is warranted (see Table~\ref{tab:iotimes}).
Our measurements show that for $1000 \times 1000$ grayscale images, Host-GPU copy times range from 100$\mu$s--200 $\mu$s, depending on whether or not we copy from page-locked memory (by setting \code{pin\_memory=True} in various places).
This Host-GPU transfer is strongly dominated by disk I/O and initial preprocessing, which is common to both CPU and GPU implementations.
The bulk of the preprocessing time is consumed by the contrast stretching operation, whose purpose is to approximately fix the threshold at which we reject spurious maxima in scale-space.
We further note that since most of our samples have ${\sim}1000$ blobs, GPU-Host copy time (i.e. ${\sim}1000$ tuples of $(x,y,\sigma)$) is nominal.
Therefore
% we argue that
Host-GPU and GPU-Host copy times are not onerous and do not dilute the performance improvements of our GPU implementation.

% \input{figures/precision_recall}
% \input{figures/multiple_gpus}

% copy times
% pin memory: True average time: 9.963878740866979e-05
% pin memory: False average time: 0.00020993370186499875

% io times
% disk read times 0.01711366940289736
% disk read+transform times 0.10486034974455834
% transform times 0.08868059366941453
% sans transform sans truth times 0.016870787553489208

% preprocess times
% img_as_float times 0.0015975497663021088
% stretch times 0.0692903277464211
% gaussian times 0.012933821510523557
% torch times 0.0003763686865568161

% stretch times 0.05872225249186158
% gaussian times 0.013208144158124924
% img_as_float times 0.001375693827867508
% torch times 0.0003293121233582497

\input{figures/gpu_vs_cpu}
\input{figures/gpu_vs_gpu}

% \ryan{We should mention MDML and NRTO somewhere as how this stuff gets used.}

We have deployed our GPU implementation of DoG as an
on-demand analysis tool through the MDML platform.
The MDML uses funcX~\cite{chard2020funcx} to serve the
DoG tool on a HPC GPU-enhanced cluster at Argonne National Laboratory.
The GPU cluster shares a high performance network with the FSP instrument
(284$\mu$s average latency).
Our DoG implementation requires 100ms on average to analyze a
PLIF image. This processing rate is sufficient to provide online
analysis of PLIF data as they are generated, enabling near real-time
feedback and optimization of configurations.

\section{Related work}\label{sec:related}

There is much historical work in this area employing classical methods and some more recent work employing neural network methods.
A useful survey is Ilonen et al.~\cite{Ilonen2018}.
Yuen et al.~\cite{Yuen1990} compare various implementations of the Hough Transform for circle finding in metallurgical applications.
Strokina et al.~\cite{Strokina2016} detect bubbles for the purpose of measuring gas volume in pulp suspensions, formulating the problem as the detection of concentric circular arrangements by using Random Sample and Consensus (RANSAC) on edges.
Poletaev et al.~\cite{Poletaev_2016} train a CNN to perform object detection for studying  two-phase bubble flows.
The primary impediment to applying Poletaev's techniques to our problem is our complete lack of ground-truthed samples, which prevents us from being able to actually train any learning machine, let alone a sample-inefficient machine such as a CNN~\cite{arora2019fine1}.
All of these methods, amongst other drawbacks, fail to be performant enough for real-time use.
For example, Poletaev's CNN takes ${\sim}$8 seconds to achieve 94\%--96\% accuracy.
One method that merits further investigation is direct estimation of the volume distribution from the power spectrum of the image~\cite{Ilonen2014}.

As pertaining to real-time optimization and steering, Laszewski et al.~\cite{osti_752879} and Bicer et al.~\cite{8109123} demonstrate real-time processing and consequent steering of synchrotron light sources.
Steering using other compute node types such as FPGAs has also been studied~\cite{7111386}.
Vogelgesang et al.~\cite{8069895} used FPGAs in concert with GPUs in streaming mode to analyze synchrotron data.
The key difference between Vogelgesang's and our work is we do not need to use direct memory access to achieve low-latency results.

\section{Conclusion}\label{sec:conclusion}

We have presented a method for the inference of droplet size distribution in PLIF images of FSP.
We briefly introduced the scale-space representation of an image and discussed several detectors that operate on such a representation.
Identifying the DoG operator as striking the right balance between accuracy and performance, we implemented it on GPU.
We then compared our GPU implementation (and a refinement thereof) against a reference implementation.
Our comparison demonstrates an order of magnitude improvement over the reference implementation with almost no decrease in accuracy. These improvements make
it possible to perform online analysis of PLIF images using a GPU-enhanced cluster,
% \ian{maybe you discuss, but what size? And, what is the difference between an HPC GPU cluster and a regular GPU cluster?}
% \maxx{I don't know what this refers to? Possibly Aurora (iirc that's the imminent HPC GPU thing at Argonne?)},
% RC: fixed
enabling online feedback and optimization of the FSP instrument.
Future work will focus on improving memory access times and disk I/O optimizations.

\begin{acks}
This work was supported by the U.S. Department of Energy, Office of Science, under contract DE-AC02-06CH11357.
% \ryan{Marius LDRD}
\end{acks}

\bibliographystyle{ACM-Reference-Format}
\bibliography{biblio}

\end{document}